\documentclass[11pt]{article}

\usepackage[final]{acl}

\usepackage{times}
\usepackage{latexsym}

\usepackage[T1]{fontenc}
\usepackage[utf8]{inputenc}

\usepackage{microtype}

\usepackage{inconsolata}

\usepackage{graphicx}

\usepackage{colortbl}
\PassOptionsToPackage{table,xcdraw}{xcolor}
\usepackage{xcolor}

\usepackage{multirow}
\usepackage{amsmath}
\usepackage{amssymb}

\usepackage{booktabs}
\usepackage{enumitem}
\usepackage{makecell}
\usepackage[ruled,longend,linesnumbered]{algorithm2e}

\newcommand{\name}{\textsc{APB-V}}

\title{\name: Accelerating Long-Video Understanding via \\Sequence-Parallelism-aware Approximate Attention}

\author{
\\
 \textbf{Yuxiang Huang\textsuperscript{1}},
 \textbf{Mingye Li\textsuperscript{2}},
 \textbf{Xu Han\textsuperscript{1}\thanks{\ \ indicates corresponding authors.}},
 \textbf{Chaojun Xiao\textsuperscript{1$*$}},
 \textbf{Weilin Zhao\textsuperscript{1}},\\
 \textbf{Ao Sun\textsuperscript{3}},
 \textbf{Ziqi Yuan\textsuperscript{1}},
 \textbf{Hao Zhou\textsuperscript{4}},
 \textbf{Fandong Meng\textsuperscript{4}},
 \textbf{Zhiyuan Liu\textsuperscript{1}}
\\
 \textsuperscript{1}{NLP Group, DCST, IAI, BNRIST, Tsinghua University, Beijing, China.}
 \\
\textsuperscript{2}{Department of CS\&T, Central South University, Changsha, China.}
\\
\textsuperscript{3}{BUPT, Beijing, China.}
\textsuperscript{4}{Pattern Recognition Center, WeChat AI, Tencent Inc.}
\\
{\tt huang-yx21@mails.tsinghua.edu.cn, \{han-xu, xcj\}@tsinghua.edu.cn}
}

\begin{document}
\maketitle
\begin{abstract}

The efficiency of long-video inference remains a critical bottleneck, mainly due to the dense computation in the prefill stage of Large Multimodal Models (LMMs).  
Existing methods either compress visual embeddings or apply sparse attention on a single GPU, yielding limited acceleration or degraded performance and restricting LMMs from handling longer, more complex videos.  
To overcome these issues, we propose \name, a sequence-parallel framework with optimized attention that accelerates long-video inference across multiple GPUs.  
By distributing approximate attention, \name~reduces computation and increases parallelism, enabling efficient processing of more visual embeddings without compression and thereby improving task performance.  
System-level optimizations, such as load balancing and fused forward passes, further unleash the potential of \name, delivering speedups of 12.72$\times$, 1.70$\times$, and 1.18$\times$ over \textsc{FlashAttn}, \textsc{ZigZagRing}, and \textsc{APB}, without notable performance loss.
Code available at \url{https://github.com/thunlp/APB}.

\end{abstract}

\section{Introduction}

The rapid advances of Large Language Models (LLMs)~\citep{achiam2023gpt, claude4, liu2024deepseek,gemini} in long-context inference have catalyzed the concurrent development of Large Multimodal Models (LMMs)~\citep{tang2023video, bai2025qwen2, zhu2025internvl3}, endowing LMMs with the capacity to understand extended video sequences. 
This capability is typically realized through the synergistic integration of visual encoders~\citep{dosovitskiy2020image} with long-context LLM backbones, yielding remarkable performance on long-video benchmarks such as LongVideoBench~\citep{wu2024longvideobench} and VNBench~\citep{zhao2024needle}.
Despite the potential in advancing long-video processing, LMMs encounter severe efficiency bottlenecks when handling ultra-long videos. 
An increased number of video frames results in a larger batch size for the visual encoder, leading to a rise in both compute cost and the number of generated video embeddings. Given the quadratic complexity of the attention layers employed by LLM backbones, these additional video embeddings extend the input length and slow down the inference speed.

\begin{figure}[t]
\begin{center}
\includegraphics[width=0.72\linewidth]{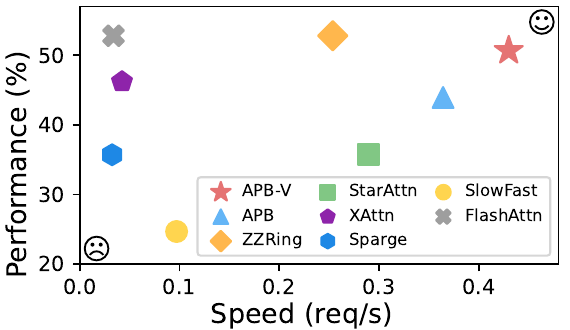}
\end{center}
\vspace{-10pt}
\caption{\name's performance and inference speed using \texttt{Qwen2.5VL-3B} as the base LMM on VNBench (processing 64-frame 1440p videos).
}
\label{fig:speed-perf}
\vspace{-10pt}
\end{figure}

% However, despite the \textit{strong potential} for advancing long-video understanding, their \textit{effectiveness} significantly degrades when processing extended video inputs.

% Despite the potential in advancing long-video understanding, LMMs encounter severe efficiency bottlenecks when handling ultra-long or high-resolution videos, with the challenges manifesting in two primary dimensions. 
% An increased number of video frames results in a larger batch size for the visual encoder, leading to a rise in both compute cost and the number of generated video embeddings. Given the quadratic complexity of the attention layers employed by LMMs, these additional video embeddings extend the input length and slow down the inference speed.

To alleviate the above efficiency issue, existing efforts can be categorized into two main approaches: \textit{\textbf{intrinsic attention optimization}} and \textit{\textbf{explicit input reduction}}.
\textit{Intrinsic attention optimization}~\citep{xu2025xattention, li2025mminference, xiao2024duoattention} methods aim to reduce the computational burden of the LLM backbone within LMMs by optimizing the attention mechanism or minimizing the key-value (KV) cache size, but cannot address the computational cost of the feed-forward networks (FFNs) and the visual encoder. 
% As a result, the effectiveness of these methods in reducing total computation and improving inference speed is limited.
\textit{Explicit input reduction} methods~\citep{xu2024slowfast, choudhury2024don, yao2025timechat} aim to reduce the number of video embeddings by applying adaptive selection or pooling operations within or after the visual encoder, thereby lowering the overall computational cost. 
% However, such reduction strategies often struggle to preserve fine-grained video details, leading to significant performance degradation, especially when the queried information is temporally or spatially sparse.
Since long-video processing typically involves computation-intensive inference, neither approach can achieve satisfactory efficiency improvements while preserving fine-grained video details.
Consequently, \textit{existing approaches inevitably face a trade-off between \textbf{efficiency} and \textbf{performance}} 
(e.g., in our experiments, the typical method \textsc{SlowFast} achieves less than $3\times$ speedup but suffers around $25\%$ accuracy degradation in Figure~\ref{fig:speed-b} and Table~\ref{tab:results-vnbench}).
% As single-GPU optimization and extreme input reduction cannot well overcome the efficiency-performance trade-off, a third scaling dimension is necessary.
As single-GPU methods usually suffer from excessive compute reduction and cannot well overcome the efficiency-performance trade-off, a third scaling dimension is necessary.

% Consequently, \textit{existing approaches inevitably face a trade-off between \textbf{efficiency} and \textbf{performance}}. In our experiment, \textsc{SlowFast} is only achieving $<3\times$ speedup (Figure \ref{fig:speed-b}) and $\sim 25\%$ performance degredation (Table~\ref{tab:results-vnbench}), this highlights the need for a third scaling dimension to sustainably advance long-video inference.
% Consequently, {\textit{accelerating long-video understanding while preserving performance remains an open problem and requires addressing the following challenges.}}

Scaling computational power while suppressing quadratic complexity, i.e., by running approximate attention with reduced compute across more GPUs (referred to as \textit{hosts}), offers a promising pathway, directly addressing the core challenges and providing the following advantages.

% Scaling computational power while suppressing the quadratic compute demand, i.e. running approximate attention with compure reduction on an increased number of GPUs (referred to as \textit{hosts}) participating in inference, offers a promising pathway, as it directly addresses the central challenges and exhibits the following advantages.

\textbf{\underline{Advantage 1:} \textbf{Supporting strong compute density through well designed parallelism.}}  
Long-video inference lends itself naturally to parallelization when guided by effective design. The video encoding process is inherently independent across frames, allowing distribution across multiple hosts. Meanwhile, the LLM backbone can be parallelized via sequence parallelism, with context segments placed and processed on different hosts. 
However, the dense computation and communication in exact sequence parallelism algorithms hinder scalability to larger host counts, resulting in limited efficiency gain (\textsc{ZZRing} in Figure~\ref{fig:speed-perf}).  
In contrast, approximate attention improves scalability through optimized computation and compressed communication, enabling faster inference.
% However, the dense compute and communication in accurate sequence parallelism algorithms limit the scalability to more hosts. 
% On the contrary, approximate attention ensures the scalability via optimized attention compute and compressed communication, thereby delivering a faster inference speed.
% Furthermore, communication-optimized approximate attention reduces attention overhead while simultaneously leveraging additional compute resources, thereby delivering faster inference speed.

% Since long videos result in heavy computational overhead for both the visual encoder and the LLM backbone, existing methods that operate after the visual encoder often overlook the computational overhead of the visual encoding process itself and thus offer only limited overall speedup. 
% % vit 需要并行
% In addition, single-GPU attention optimization often fails to sufficiently reduce the cost of LLM backbone to meet demands, as the compute bandwidth provided is strictly limited. 
% % 需要稀疏attn
% Therefore, a promising approach is to utilize more GPUs to enhance compute capacity and apply approximate attention to simultaneously reduce computational overhead.
% % Based on this, we initiate our optimization by employing frame-level parallelism for the visual encoder and sequence parallelism for the LLM backbone, integrated with approximate attention to further reduce the overall compute load.

\textbf{\underline{Advantage 2:} \textbf{Preserving task performance.}}  
Conventional optimizations often sacrifice fine-grained details in long-video inputs, especially when token pruning is applied for explicit input reduction. 
% By introducing greater computational power and leveraging carefully designed approximate attention mechanisms, 
By replacing input reduction with approximate attention compute reduction, the full set of visual embeddings is preserved, thereby safeguarding task performance.

% video encoding and context processing can run in parallel without compressing visual embeddings, thus preserving video details and task performance.

% To address the these challenges, we propose \name, a sequence parallelism-aware approximate attention framework designed to accelerate long-video understanding. Our main contributions are summarized as follows:

Building on this perspective, we propose \name{}.
Our main contributions are summarized as follows:

\textbf{\underline{(I)}} We introduce \name, a sequence-parallelism-aware approximate attention framework for accelerating long-video understanding. 
Unlike embedding compression or single-GPU attention optimization, \name~adopts local KV cache compression with a distributed approximate attention mechanism, well striking an efficiency-performance balance.

\textbf{\underline{(II)}} We optimize \name~from a system-level perspective by designing a load-balancing strategy across hosts, along with a two-stage attention mechanism where communication perfectly overlapped with computation.

\textbf{\underline{(III)}} We evaluate \name~on extensive benchmarks. As shown in Table~\ref{fig:speed-perf}, the experimental results show that \name~exhibits the best tradeoff between performance and efficiency. More specifically, \name~achieves speedup of 12.72$\times$, 1.70$\times$, and 1.18$\times$ over \textsc{FlashAttn}, \textsc{ZigZagRing}, and \textsc{APB}, without significant performance degredation. 
% These baselines are typical methods for long-video understanding.
% \end{itemize}

\begin{figure*}[t]
\begin{center}
\includegraphics[width=1\linewidth]{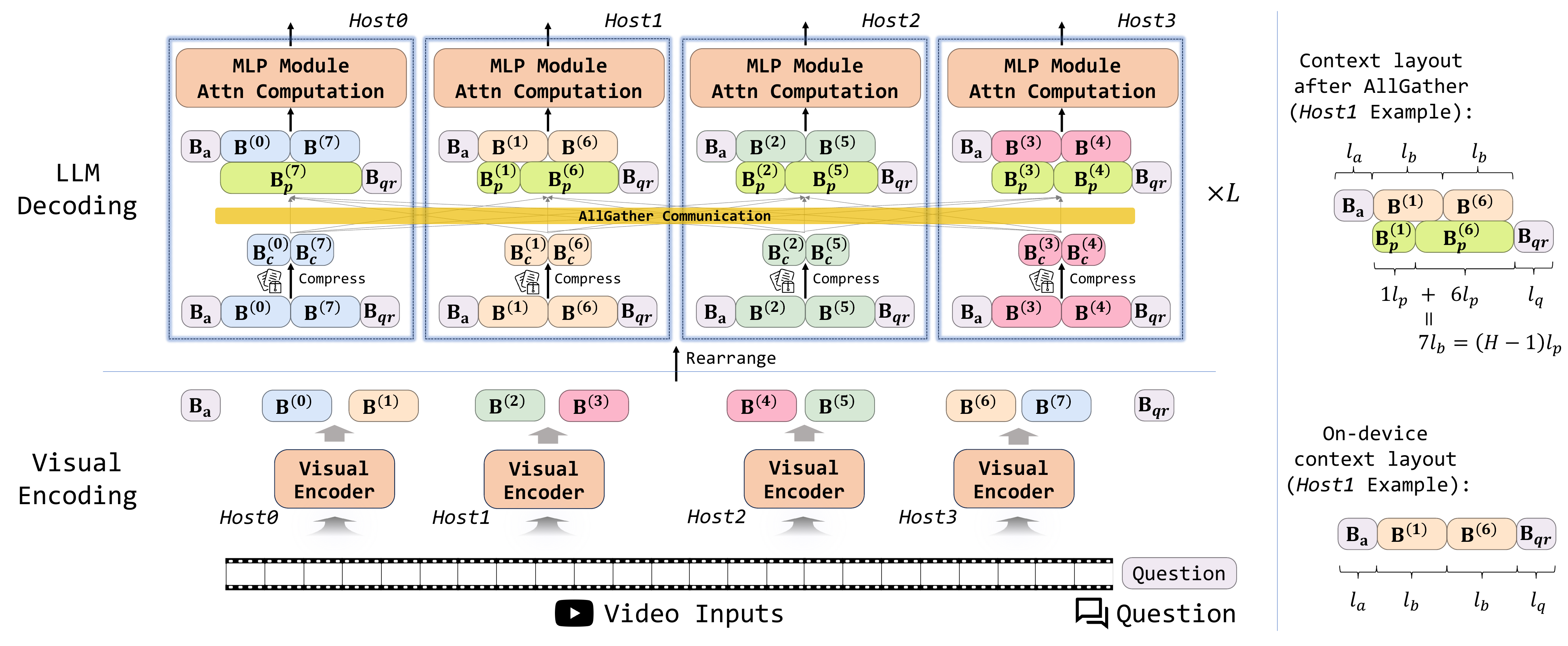}
\end{center}
\vspace{-10pt}
\caption{The framework of \name. The anchor block and passing block are denoted as $\mathbf{B}_a$ and $\mathbf{B}_p$, while $\mathbf{B}^{(h)}$ denotes the context block on virtual host $h$. $\mathbf{B}_{qr}$ represents the query block.
The video input is first encoded into embeddings using frame parallelism across hosts. After context splitting, each physical host (containing two virtual hosts) holds the anchor block, query block, and corresponding context blocks. In approximate attention, each context block is first compressed then communicated. Attention is then computed over $\mathbf{B}_a$, $\mathbf{B}^{(h)}$, $\mathbf{B}^{(2H-1-h)}$, $\mathbf{B}_p^{(h)}$, $\mathbf{B}_p^{(2H-1-h)}$, and $\mathbf{B}_{qr}$. The passing blocks are discarded immediately after attention.
}
\label{fig:framework}
\vspace{-10pt}
\end{figure*}

% \section{Methodology}

% As shown in Figure~\ref{fig:framework}, we propose \name to accelerate long-video understanding, which adopts frame-level parallelism in the visual encoder and load-balanced sequence-parallelism-aware approximate attention in the LLM backbone. 
% In this section, by first introducing preliminaries and notations, we will present the inference process of \name, and then provide a detailed description of system optimizations designed to further enhance inference speed. Figure~\ref{fig:framework} shows the whole framework of \name.

\section{Preliminaries}
\label{seq:pre}
% \subsection{Preliminaries and Notations}
We introduce the basic notations of Transformers and long-video inference in this section.

\textbf{Transformers.} 
For an $L$-layer Transformer-based model, each layer consists of an attention block and an FFN block~\citep{vaswani2017attention}. The model takes a sequence of $d$-dimensional embeddings $\mathbf{E} \in \mathbb{R}^{n \times d}$ as the input, where $n$ is the input sequence length. For each layer, given the input hidden $\mathbf{H}$, the query, key, and value matrices are first computed as 
$
\mathbf{Q} = \mathbf{H}\mathbf{W}_Q,
\mathbf{K} = \mathbf{H}\mathbf{W}_K,
\mathbf{V} = \mathbf{H}\mathbf{W}_V,
$
and then the attention is applied as
\begin{equation}
% \scalebox{0.9}{
\begin{aligned}
    \mathbf{A} &= \texttt{Attn}\left(\mathbf{Q}, \mathbf{K}, \mathbf{V}\right) \\
    &=\texttt{Softmax}\left(\frac{\mathbf{Q}\mathbf{K}^{\top} \odot \mathbf{M}}{\sqrt{d}}\right)\cdot \mathbf{V},
\end{aligned}%}
\end{equation}
where $\mathbf{M}$ is the attention mask. Based on the attention block, the FFN block computation is
\begin{equation}
% \scalebox{0.9}{
\begin{aligned}
    \mathbf{H}' = \texttt{FFN}\left(\mathbf{A}\right).
\end{aligned}%}
\end{equation}
We omit the notations for attention heads, position embeddings, and normalizations for simplicity.

\textbf{Long-Video Inference.} An LMM $\mathcal{M}$ typically consists of a visual encoder $\mathcal{M}_v$, a connector $\mathcal{M}_c$, an embedding layer $\mathcal{M}_e$, and an LLM backbone $\mathcal{M}_l$. During the long-video inference, the video is first divided into $F$ frames, each containing $H \times W$ $d_v$-dimensional patches. We denote the patchified input as a fourth-order tensor $\mathbf{I} \in \mathbb{R}^{F \times H \times W \times d_v}$. For the $i$-th frame $\mathbf{I}_i$, it is first reshaped into a second-order embedding sequence $\tilde{\mathbf{I}}_i \in \mathbb{R}^{HW \times d_v}$, and then encoded by the visual encoder, producing the output $\mathbf{O}_i = \mathcal{M}_v(\tilde{\mathbf{I}}_i) \in \mathbb{R}^{HW \times d_v}$. 
By encoding each frame output $\mathbf{O}_i$ to $n_v$ $d$-dimensional embeddings through the connector $\mathcal{M}_c$, we get the whole video embeddings $\mathbf{E}_v = \mathcal{M}_c\left([\mathbf{O}_1,\ldots,\mathbf{O}_F]\right) \in \mathbb{R}^{Fn_v \times d}$. Finally, the video embeddings and the text embeddings (encoded from the textual context by $\mathcal{M}_e$) are concatenated to form the input for the LLM $\mathcal{M}_l$, denoted as $\mathbf{E} = [\mathbf{E}_v, \mathbf{E}_t] \in \mathbb{R}^{n \times d}$, where $\mathbf{E}_t \in \mathbb{R}^{n_t \times d}$ is the text embeddings and $n = n_v + n_t$.

\section{\name's Framework Design}

\name~is a long-video inference framework that leverages specially designed sequence parallelism and approximate attention, as shown in Figure~\ref{fig:framework}. In this section, we present the main components in the order they operate during inference.
We assume the input to an LMM $\mathcal{M}$ is $[\mathbf{I}, \mathbf{T}_{\text{Q}}]$, where $\mathbf{T}_{\text{Q}}$ denotes the token sequences of the query, and $\mathbf{I}$ represents the patchified video input. 
We define a \textbf{\textit{host}} as a group of GPUs that maintains a full replica of the LMM. 
Inference is performed across $H$ such hosts.

\subsection{Frame Parallelism}

Initially, an LMM's visual encoder transforms the visual inputs into embeddings, as shown in Section~\ref{seq:pre}.
Since each frame is encoded independently, we adopt frame parallelism to address the high computational intensity, inspired by \textsc{LongVILA}~\citep{chen2024longvila}.
On the $h$-th host, we obtain the input embeddings of the entire text query and a subset of video frames as
\begin{equation}
% \scalebox{0.9}{
\begin{aligned}
    \mathbf{E}_v^{(h)}, \mathbf{E}_{Q} = \mathcal{M}_c\left(\mathcal{M}_v\left(\mathbf{I}^{(h)}\right)\right), \mathcal{M}_e\left(\mathbf{T}_{\text{Q}}\right),
\end{aligned}%}
\end{equation}
where $\mathbf{I}^{(h)}$ is a subset of $\mathbf{I}$ for the $h$-th host.

% At the beginning, the token and visual inputs are transformed into input embeddings. 
% Given the high computational intensity of the visual encoder, we adopt frame parallelism by evenly distributing video frames across all hosts, inspired by \textsc{LongVILA}~\citep{chen2024longvila}, without inter-host communication. 
% % Since the visual encoding process is independent across frames, this parallelism eliminates the need for inter-host communication.
% On each host $h$, we obtain the input embeddings of all token sequences and a subset of video frames, formalized as
% \begin{equation}
% \scalebox{0.9}{
% \begin{aligned}
%     \mathbf{E}_v^{(h)}, \mathbf{E}_{Q} = \mathcal{M}_c\left(\mathcal{M}_v(\mathbf{I}^{(h)})\right), \mathcal{M}_e\left(\mathbf{T}_{\text{Q}}\right).
% \end{aligned}}
% \end{equation}

\subsection{Context Splitting}

% Since the coarse-grained frame structure may lead to load imbalance, we perform a communication to enable each host to obtain a complete copy of all input embeddings by $\mathbf{E}_v = \texttt{AllGather}\left(\{\mathbf{E}_v^{(h)}\}_{h=1}^{H}\right)$. Then, the input embeddings of each host are partitioned into three components: 
We begin with a collective communication step that enables each host to access the complete sequence: $\mathbf{E}_v = \texttt{AllGather}\left(\{\mathbf{E}_v^{(h)}\}_{h=1}^{H}\right)$. 
Then, the input embeddings are partitioned into three components:
\begin{itemize}[leftmargin=*, noitemsep, nolistsep]
    \item \textbf{\textit{Anchor Block}}: $\mathbf{B}_a = [e_1, \cdots, e_{l_a}]$, comprising the initial $l_a$ embeddings of $\mathbf{E}_v$;
    \item \textbf{\textit{Query Block}}: $\mathbf{B}_{qr} = \mathbf{E}_{Q}$, consisting of all query embeddings at the end of the sequence;
    \item \textbf{\textit{Context Block}}: $\mathbf{B}^{(h)}$, formed by equally dividing the remaining sequence of $\mathbf{E}_v$ across hosts. 
\end{itemize}
% the \textbf{\textit{anchor block}} $\mathbf{A} = [e_1, \cdots, e_{l_a}]$ comprising the initial $l_a$ embeddings of $\mathbf{E}_v$, 
% the \textbf{\textit{query block}} $\mathbf{B}_{qr} = \mathbf{E}_{Q}$ consisting query embeddings, 
% and the \textbf{\textit{context block}} $\mathbf{B}_h$ formed by equally dividing the remaining sequence of $\mathbf{E}_v$ across hosts. 
% Unlike previous methods, \name~enforces non-overlap between the anchor block and all context blocks. 
% Considering the causal nature of Transformer-based models, i.e., the computational cost of those blocks at the end of the sequence is greater than that of the blocks at the front of the sequence.
% However, the vanilla design of \textsc{APB} suffers from load imbalance, as different hosts process passing blocks of varying lengths.

In decoder-only models, different context blocks attend to varying amounts of context, and naive host assignment thus leads to imbalanced attention computation.
We introduce a ZigZag-style arrangement to solve this issue.
% To address this issue and further improve the prefilling speed, we introduce a ZigZag-style attention mechanism. 
Specifically, we introduce $2H$ virtual hosts during context splitting. A \textbf{\textit{virtual host}} is a logical construct that does not correspond directly to any physical GPU assignment, whereas a \textbf{\textit{physical host}} corresponds to an actual set of GPUs. Each physical host contains two complementary virtual hosts to ensure better load balancing.
More details of the mapping from physical to virtual hosts are introduced in Section~\ref{sec:zigzag}.
For simplicity in notations, we use virtual hosts to describe the following attention process.
% Details on mapping virtual hosts to physical hosts are discussed in the system optimization section, and we introduce the following algorithmic inference process from the perspective of virtual hosts.
% Note that, since the anchor block and the query block are shared across all virtual hosts, only a single copy is maintained on each physical host to reduce memory usage and redundant computation. Detailed load balancing strategies are discussed in the system optimization section.

\subsection{Approximate Attention}

To reduce both the computational and communication overhead of attention, we design an approximate attention mechanism based on the observation that only the most essential KV pairs need to be visible to subsequent tokens, whereas the others can remain confined within their local blocks, avoiding unnecessary cross-host attention.

The $h$-th virtual host maintains the anchor block $\mathbf{B}_a$, the context block $\mathbf{B}^{(h)}$, and the query block $\mathbf{B}_{qr}$, and the QKV states of these blocks are $\{\mathbf{Q}_a, \mathbf{K}_a, \mathbf{V}_a\}$, $\{\mathbf{Q}^{(h)}, \mathbf{K}^{(h)}, \mathbf{V}^{(h)}\}$, $\{\mathbf{Q}_{qr}, \mathbf{K}_{qr}, \mathbf{V}_{qr}\}$, respectively.
% First, we perform KV cache compression by leveraging the query block to identify the most $l_p$ important KV pairs via attention scores: 
On the $h$-th host, we first use the query-to-context attention scores $\mathbf{Q}_{qr}\mathbf{K}^{(h)\top}$ to identify the most $l_p$ important KV pairs $(\mathbf{K}_c^{(h)}, \mathbf{V}_c^{(h)})$ from $(\mathbf{K}^{(h)}, \mathbf{V}^{(h)})$.
% $\texttt{idx} = \text{argtop}_{l_p}\left(\mathbf{Q}^{qr}\mathbf{K}_h^\top\right)$, and $\mathbf{K}_h^C, \mathbf{V}_h^C = \mathbf{K}_h[\texttt{idx}], \mathbf{V}_h[\texttt{idx}]$.
Then, we build the \textbf{\textit{passing blocks}} that contain the essential KV pairs on previous hosts whose host index is no larger than $h$, similar to \textsc{APB}~\citep{huang2025apb}, by
$\mathbf{K}_p^{(h)}, \mathbf{V}_p^{(h)} = \texttt{AllGather}\left(\{\mathbf{K}_c^{(h)}, \mathbf{V}_c^{(h)}\}_{h=1}^{H}\right)_{\leq h}$.
Finally, we compute the attention scores for the anchor and context blocks:
% \scalebox{0.83}{
\begin{equation}
\scalebox{0.83}{
\ensuremath{
    \begin{aligned}
        \mathbf{A}_a &= \texttt{Attn}\left(\mathbf{Q}_a, \mathbf{K}_a, \mathbf{V}_a\right), \\
        \mathbf{A}_h &= \texttt{Attn}\left(\mathbf{Q}^{(h)}, \left[\mathbf{K}_a,\mathbf{K}_p^{(h)},\mathbf{K}^{(h)}\right], \left[\mathbf{V}_a,\mathbf{V}_p^{(h)},\mathbf{V}^{(h)}\right]\right),
    \end{aligned}
    }}
\end{equation}
where $\mathbf{A}_a, \mathbf{A}^{(h)}$ correspond to the attention results of the anchor and context blocks.

We simultaneously compute the attention result $\mathbf{A}^{qr}$ for the query block, utilizing \textsc{FlashAttn}'s \texttt{lse} to merge the results from different hosts:
\begin{equation}
\scalebox{0.9}{\ensuremath{
    \begin{aligned}
    &\mathbf{A}_{qr}^{(h)}, \texttt{lse}^{(h)} = \texttt{Attn}(\mathbf{Q}_{qr}, \mathbf{\tilde K}_{qr}, \mathbf{\tilde V}_{qr}), \\
    &\mathbf{A}_{qr} = \texttt{Merge}(\texttt{AllGather}(\{\mathbf{A}_{qr}^{(h)}, \texttt{lse}^{(h)})\}_{h=1}^H),
    \end{aligned}}}
\end{equation}
where $\mathbf{\tilde K}_{qr} = [\mathbf{K}_{a}^{(h)}, \mathbf{K}^{(h)}, \mathbf{K}_{qr}]$, with $\mathbf{K}_{a}^{(h)}$ denoting the sliced anchor block for load balancing such that $[\mathbf{K}_{a}^{(h)}]_{h=1}^H = \mathbf{K}_a$. (The same applies to $\mathbf{\tilde V}_{qr}$.)
Further details are provided in Appendix~\ref{sec:pseudocode}.

% 这段放到system optimiztation里
% Note that each physical host contains two virtual hosts. To avoid redundant computation, we do not recompute the anchor block when processing the second virtual host. Moreover, the query attention is computed jointly over both context blocks associated with the two virtual hosts. Further details are provided in the system optimization section.

\begin{table*}[t]
\small
\centering
\scalebox{0.9}{
\begin{tabular}{l|cccc|cccc|cccc|c}
\toprule
\multirow{2}{*}{Method} & \multicolumn{4}{c|}{Retrieval} & \multicolumn{4}{c|}{Ordering} & \multicolumn{4}{c|}{Counting} & \multirow{2}{*}{Overall}\\

 & E & I-1 & I-2 & Avg. & E & I-1 & I-2 & Avg. & E-1 & E-2 & I & Avg. &\\
\midrule
\multicolumn{14}{c}{\texttt{InternVL3-2B}} \\ \midrule
\textsc{FullAttn} & 90.00 & 90.67 & 36.00 & 72.22 & 64.67 & 24.00 & 24.67 & 37.78 & 40.67 & 4.67 & 28.67 & 24.67 & 44.89
\\ \midrule
\textsc{XAttn} & 90.00 & \textbf{90.67} & \textbf{38.00} & \textbf{72.89} & 54.00 & \textbf{23.33} & 15.33 & 30.89 & \textbf{37.33} & \textbf{7.33} & \textbf{28.67} & \textbf{24.44} & 42.74
\\
\textsc{Sparge} & \textbf{91.33} & 89.33 & 33.33 & 71.33 & 46.67 & 11.33 & 18.67 & 25.56 & 26.67 & 4.67 & 24.67 & 18.67 & 38.52
\\
\textsc{SlowFast} & 48.00 & 56.67 & 32.00 & 45.56 & 15.33 & 6.67 & 8.67 & 10.22 & 24.67 & 5.33 & 26.00 & 18.67 & 24.81
\\
\textsc{StarAttn} & 90.00 & 88.67 & 34.67 & 71.11 & 29.33 & 6.00 & 10.00 & 15.11 & 18.00 & 6.00 & 22.00 & 15.33 & 33.85
\\
\textsc{APB} & 89.33 & 89.33 & 36.00 & 71.56 & 59.33 & 17.33 & 18.00 & 31.56 & 33.33 & 3.33 & 24.00 & 20.22 & 41.11
\\
\rowcolor{pink!20}
\textbf{\name} & {90.67} & 89.33 & 32.67 & 70.89 & \textbf{64.00} & 22.00 & \textbf{21.33} & \textbf{35.78} & \textbf{37.33} & 5.33 & 26.67 & 23.11 & \textbf{43.26}
\\
\midrule
\multicolumn{14}{c}{\texttt{Qwen2.5VL-3B}} \\ \midrule
\textsc{FullAttn} & 90.67 & 84.00 & 48.00 & 74.22 & 72.67 & 51.33 & 37.33 & 53.78 & 53.33 & 8.00 & 30.00 & 30.44 & 52.81
\\ \midrule
\textsc{XAttn} & \textbf{90.00} & 77.33 & \textbf{48.00} & 71.78 & 65.33 & 39.33 & 23.33 & 42.67 & 36.67 & \textbf{10.00} & 26.00 & 24.22 & 46.22
\\
\textsc{Sparge} &
89.33 & 75.33 & 37.33 & 67.33 & 36.67 & 13.33 & 10.00 & 20.00 & 26.00 & 8.00 & 25.33 & 19.78 & 35.70
\\
\textsc{SlowFast} & 41.33 & 64.00 & 25.33 & 43.56 & 12.00 & 10.00 & 12.67 & 11.56 & 23.33 & 5.33 & 28.00 & 18.89 & 24.67
\\
\textsc{StarAttn} & \textbf{90.00} & 83.33 & 46.67 & \textbf{73.33} & 18.67 & 10.67 & 10.67 & 13.33 & 27.33 & 6.67 & 27.33 & 20.44 & 35.70
\\
\textsc{APB} & \textbf{90.00} & \textbf{82.67} & 44.00 & 72.22 & 47.33 & 29.33 & 22.00 & 32.89 & 44.67 & 8.00 & 27.33 & 26.67 & 43.93
\\
\rowcolor{pink!20}
\textbf{\name} & \textbf{90.00} & \textbf{82.67} & 46.67 & {73.11} & \textbf{66.00} & \textbf{47.33} & \textbf{33.33} & \textbf{48.89} & \textbf{52.00} & 9.33 & \textbf{28.67} & \textbf{30.00} & \textbf{50.67}
\\
\midrule
\multicolumn{14}{c}{\texttt{Qwen2.5VL-7B}} \\ \midrule
\textsc{FullAttn} & 90.67 & 82.00 & 59.33 & 77.33 & 74.67 & 59.33 & 57.33 & 63.78 & 54.67 & 13.33 & 34.67 & 34.22 & 58.44
\\ \midrule
\textsc{XAttn} & \textbf{90.67} & 79.33 & \textbf{60.00} & \textbf{76.67} & 70.67 & 53.33 & 44.67 & 56.22 & 47.33 & 12.00 & \textbf{32.00} & 30.44 & 54.44
\\
\textsc{Sparge} & 88.67 & 79.33 & 50.67 & 72.89 & 63.33 & 35.33 & 38.00 & 45.56 & 44.00 & 10.67 & 26.67 & 27.11 & 48.52
\\
\textsc{SlowFast} & 43.33 & 69.33 & 42.67 & 51.78 & 14.00 & 12.00 & 9.33 & 11.78 & 22.00 & 9.33 & 28.67 & 20.00 & 27.85
\\
\textsc{StarAttn} & \textbf{90.67} & \textbf{82.00} & 57.33 & \textbf{76.67} & 27.33 & 20.00 & 18.67 & 22.00 & 25.33 & 12.00 & 28.67 & 22.00 & 40.22
\\
\textsc{APB} & 89.33 & \textbf{82.00} & 58.67 & \textbf{76.67} & 58.00 & 35.33 & 36.67 & 43.33 & 48.00 & 11.33 & 30.00 & 29.78 & 49.93
\\
\rowcolor{pink!20}
\textbf{\name} & \textbf{90.67} & 80.67 & 58.00 & 76.44 & \textbf{72.00} & \textbf{56.67} & \textbf{48.00} & \textbf{58.89} & \textbf{54.00} & \textbf{14.00} & \textbf{32.00} & \textbf{33.33} & \textbf{56.22} 
\\
\bottomrule
\end{tabular}}

\caption{Accuracies (\%) of VNBench. 
% Higher scores indicate better performance, and ``Overall'' stands for the overall accuracy on the complete dataset. 
``E'' and ``I'' represent the edited and inserted data subset, respectively.
% The highest score in each column is marked in \textbf{bold}.
}
\label{tab:results-vnbench}
\end{table*}

\section{\name's System Optimizations}

\begin{figure}[t]
% \vspace{-10pt}
\begin{center}
\includegraphics[width=\linewidth]{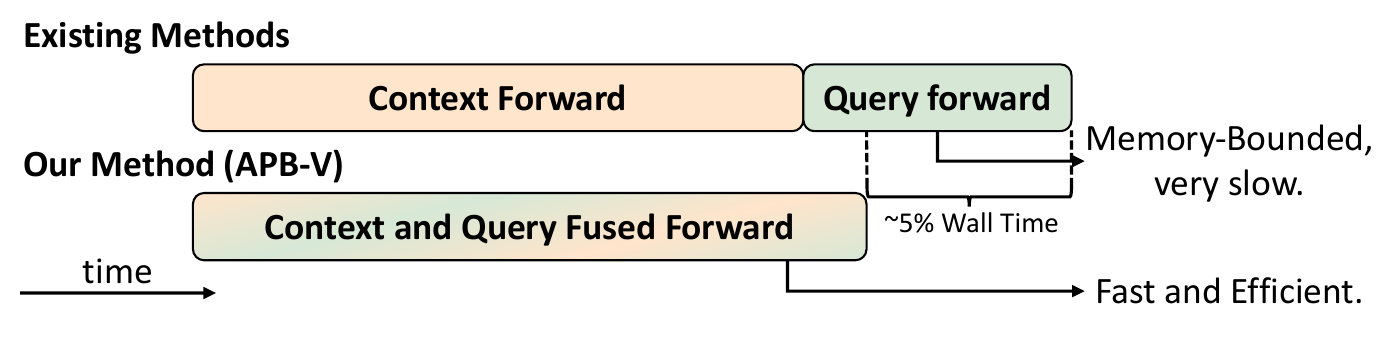}
\end{center}
\vspace{-12.5pt}
\caption{Fused context and query forward pass.}
\vspace{-8pt}
\label{fig:fused}
\end{figure}
\begin{figure}[t]

\begin{center}
\includegraphics[width=\linewidth]{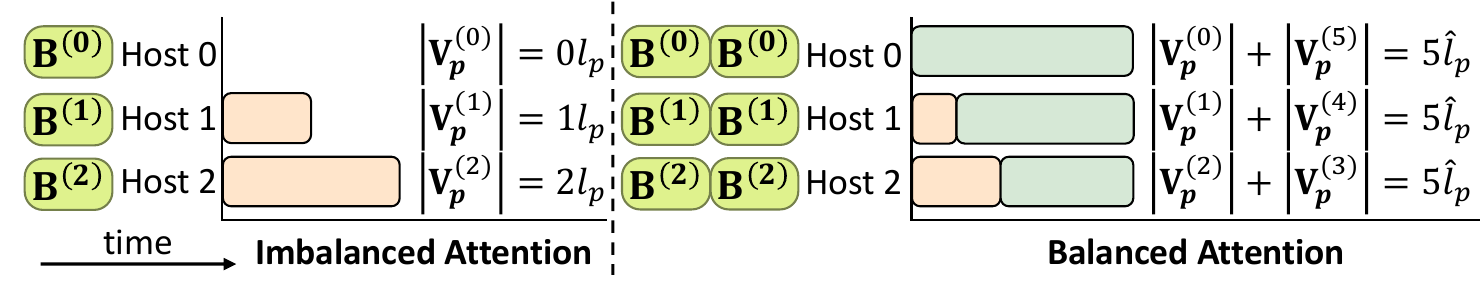}
\end{center}
\vspace{-12.5pt}
\caption{Attention load balancing. ``$|\mathbf{V}_c^{(h)}|$'' is the total length of previous essential KVs of virtual host $h$.}
\vspace{-10pt}
\label{fig:balance}
\end{figure}
\begin{figure}[t]
% \vspace{-10pt}
\begin{center}
\includegraphics[width=0.9\linewidth]{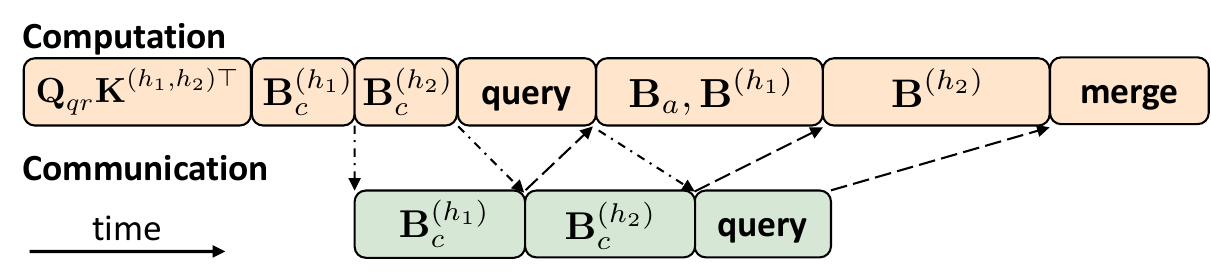}
\end{center}
\vspace{-10pt}
\caption{Overlapping communication with computation on virtual host $h$.  
``$\mathbf{Q}_{qr}\mathbf{K}^{(h_1,h_2)\top}$'' estimates KV importance; ``$\mathbf{B}_{c}^{(h_1,h_2)}$'' are essential KVs; ``\texttt{query}'' and ``\texttt{merge}'' denote query attention and its merging; ``$\mathbf{B}_a$'' and ``$\mathbf{B}^{(h_1,h_2)}$'' indicate anchor and context attention.}
\vspace{-10pt}
\label{fig:overlap}
\end{figure}

The efficiency in existing methods~\citep{acharya2024starattention, huang2025apb} that adopt sequence-parallelism inference is usually bounded by:
\begin{itemize}[leftmargin=*, noitemsep, nolistsep]
    \item Imbalanced video encoding across hosts;
    \item Memory-bound forward pass for query prefill;
    \item Imbalanced attention computation across hosts;
    \item Inefficient communication design.
\end{itemize}
We briefly introduce the following system optimizations to alleviate these obstacles, with more details introduced in Appendix~\ref{sec:sys_opt}.

\subsection{Visual Load Balancing} 

To enable balanced frame parallelism in the visual encoder, we design a visual load-balancing strategy by evenly distributing the number of frames across virtual hosts. 
Given $F$ frames and $H$ virtual hosts, the number of frames assigned to the $h$-th host is:
\begin{equation}
\scalebox{0.9}{\ensuremath{
\begin{aligned}
    F^{(h)} = \left\lfloor \frac{F}{H} \right\rfloor + \mathbb{I}\left[h < F \bmod H\right],
\end{aligned}}}
\end{equation}
where $\mathbb{I}$ denotes the indicator function.

\subsection{Fused Context and Query Forward}

Previous methods execute the prefill stage for context and query separately.  
Since queries are typically short, a dedicated forward pass for the query often becomes memory-bound and inefficient, as shown in Figure \ref{fig:fused}.  
To overcome this, we fuse context and query blocks into a single forward pass, computing attention jointly.  
The partial query result $\mathbf{A}_{qr}^{(h)}$ is merged using the $\texttt{lse}$ output from \textsc{FlashAttn} at the end of each attention layer.

\subsection{Approximate Attention Load Balancing}
\label{sec:zigzag}

Adopting vanilla sequence parallelism for approximate attention incurs imbalanced attention computation: for host $h$, the number of passing blocks is $h \cdot l_p$, leading to uneven workloads.
% \textsc{APB}~\citep{huang2025apb} introduces imbalanced computation across hosts, as the number of context blocks involved in the attention computation varies from host to host. This imbalance limits overall efficiency, since the attention computation time is ultimately bounded by the last host.
To mitigate this, we adopt a ZigZag-style~\citep{zhu2024ringflash} load-balancing strategy, assigning the $h$-th and $(2H-1-h)$-th virtual hosts to the same physical host, so that the total length of passing blocks is balanced across hosts, as shown in Figure \ref{fig:balance}.

% Given $H$ physical hosts, we instantiate $2H$ virtual hosts and assign virtual hosts $h$ and $2H-1-h$ to physical host $h$. 
% Since anchor blocks are identical, each host only holds one replica of the anchor block, and so is the query block.
% In this setup, each host processes one anchor block of length $l_a$, two context blocks of length $l_b$, and a total of $2H - 1$ passing blocks of length $l_p$, ensuring an $h$-independent and equal amount of compute for all physical hosts. 

% Since the anchor block introduces a significant amount of compute in the query attention, we also design a load balancing strategy specifically for this stage. As each physical host holds a full copy of the anchor block, we evenly divide the anchor block into $H$ slices. The query on physical host $h$ attends only to the $h$-th slice of the anchor block. 
% The numerical accuracy is ensured by the online \texttt{softmax} technique.

% TODO: 加张图

\subsection{Overlapped Communication}

Existing methods like \textsc{StarAttn}~\cite{acharya2024starattention} avoid inter-host communication but fail to capture long-term dependencies.  
In contrast, \textsc{APB}~\citep{huang2025apb} incorporates inter-host communication to model such dependencies, at the cost of higher communication overhead.
% as attention computation becomes tightly coupled with communication results.
To address this, we overlap communication with computation, as illustrated in Figure~\ref{fig:overlap}.  
The transfer of passing blocks and partial query attention results are performed concurrently with attention calculation.

\section{Experiments}

We now present our empirical analysis of \name, focusing on the following questions.

\noindent(\textbf{\textit{Q1}}) Can \name~achieve similar or better task performance compared with other baselines? 

\noindent(\textbf{\textit{Q2}}) Can \name~obtain a faster inference speed under various video lengths and resolutions? 

\noindent(\textbf{\textit{Q3}}) How does each component of \name~contribute to overall performance and efficiency?

We also provide a case study in Section~\ref{sec:case}.

\subsection{Experimental Settings}

\textbf{Benchmarks.} 
We evaluate two long-video benchmarks: LongVideoBench~\citep{wu2024longvideobench} and VNBench~\citep{zhao2024needle}. 
LongVideoBench contains 1337 real-world videos in four duration ranges (8s-15s, 15s-60s, 180s-600s, and 900s-3600s). VNBench offers 1350 synthetic videos featuring challenging retrieval, ordering, and counting tasks.
We set the frame number to 64 for both benchmarks.

\textbf{Models.} 
To examine the effect of \name~ on various model architectures and model sizes, we conduct experiments based on three LMMs: \texttt{InternVL3-2B}~\citep{zhu2025internvl3}, and \texttt{Qwen2.5VL-3B/7B}~\citep{bai2025qwen2}. \texttt{InternVL3-2B} resizes frames to 448$\times$448, while \texttt{Qwen2.5VL} models supports native-resolution processing using the patch size of 14$\times$14.

\textbf{Metrics.}
We use the benchmarks' original metrics for performance evaluation. For speed measurement, we define inference speed as requests processed per second (req / s) and compute relative speedup against \textsc{FullAttn}.

\textbf{Baselines.}
Here are our four baseline categories:
\begin{itemize}[leftmargin=*, noitemsep, nolistsep]
\item Accurate attention including \textsc{FullAttn}~\citep{dao2023flashattention} and \textsc{ZigZagRingAttn} (denoted as \textsc{ZZRing})~\citep{zhu2024ringflash}; 
\item Token pruning \textsc{SlowFast-Llava} (denoted as \textsc{SlowFast})~\citep{xu2024slowfast};
\item Sparse attention \textsc{XAttn}~\citep{xu2025xattention} and \textsc{Sparge}~\citep{zhang2025spargeattn};
\item Approximate attention with sequence parallelism including \textsc{StarAttn}~\citep{acharya2024starattention} and \textsc{APB}~\citep{huang2025apb}. 
\end{itemize}
For methods using sequence parallelism (\textsc{ZZRing}, \textsc{StarAttn}, and \textsc{APB}), we employ frame parallelism on the visual encoder to ensure a fair competition and set the host number to 8 with 1 GPU in each host. Other methods (\textsc{FlashAttn}, \textsc{XAttn}, \textsc{Sparge}, and \textsc{SlowFast}) use a single GPU. We train \textsc{APB}'s retaining heads to select KV blocks on NextQA~\citep{xiao2021next} with other settings identical to the official configuration.

\textbf{Hyperparameters and Environments.}
Given $n$ input embeddings, we set the anchor length $l_a = \frac{n}{64}$ and the passing length $l_p = \frac{n}{128}$ for \name, while \textsc{APB} uses $l_a = \frac{n}{64}$ and $l_p = \frac{n}{64}$. With \name's $2H$ virtual hosts, \name~and \textsc{APB} finally maintain the same amount of compute. When adapting \textsc{SlowFast} to \texttt{InternVL-2B}, we remove all the text labels of each frame; we gather the corresponding ids for 3D-RoPE~\citep{bai2025qwen2} when adapting to \texttt{Qwen2.5-VL}.
For \textsc{Sparge}, we set the threshold of similarity and cumulative distribution to 0.3 and 0.96, respectively. Other baseline methods use default configurations. All experiments are conducted on an 8$\times$A800-40GB cluster with 3rd-generation NVLink interconnects.

\begin{table}[t]
\small
\centering
\setlength{\tabcolsep}{0.8mm}
\scalebox{0.9}{
\begin{tabular}{l|cccc|cc}
\toprule
Method & 8-15s & 15-60s & 180-600s & 900-3600s & Overall & P. D.\\
\midrule
\multicolumn{7}{c}{\texttt{InternVL3-2B}} \\ \midrule
\textsc{FullAttn} & 61.90 & 67.44 & 54.61 & 50.00 & 55.35 & -
\\ \midrule
\textsc{XAttn} & 62.43 & 71.51 & 53.40 & 48.58 & 54.97 & Yes
\\
\textsc{Sparge} & 56.08 & 64.53 & 50.49 & 42.55 & 49.74 & Yes
\\
\textsc{SlowFast} & 60.85 & 66.86 & 50.24 & 44.50 & 51.46 & Yes
\\
\textsc{StarAttn} & \textbf{63.49} & 66.86 & 52.91 & 48.40 & 54.30 & Yes
\\
\textsc{APB} & 61.90 & \textbf{67.44} & \textbf{55.83} & 48.76 & 55.20 & Yes
\\
\rowcolor{pink!20}
\textbf{\name} & 62.43 & \textbf{67.44} & 54.85 & \textbf{49.82} & \textbf{55.42} & \textbf{No}
\\
\midrule
\multicolumn{7}{c}{\texttt{Qwen2.5VL-3B}} \\ \midrule
\textsc{FullAttn} & 69.31 & 70.93 & 52.18 & 43.79 & 53.47 & -
\\ \midrule
\textsc{XAttn} & 66.14 & 69.19 & 52.18 & 44.33 & 53.03 & Yes
\\
\textsc{Sparge} & 59.79	& 60.47	& 42.48	& 43.97 & 47.87 & Yes
\\
\textsc{SlowFast} & 65.61 & 64.53 & 50.49 & {46.45} & 52.73 & Yes
\\
\textsc{StarAttn} & \textbf{68.78} & 70.93 & 53.64 & 45.39 & 54.52 & \textbf{No}
\\
\textsc{APB} & 68.25 & \textbf{72.09} & 52.18 & \textbf{47.34} & \textbf{54.97} & \textbf{No}
\\
\rowcolor{pink!20}
\textbf{\name} & 68.25 & 70.93 & \textbf{53.88} & {44.86} & {54.30} & \textbf{No}
\\
\midrule
\multicolumn{7}{c}{\texttt{Qwen2.5VL-7B}} \\ \midrule
\textsc{FullAttn} & 73.81 & 73.84 & 56.44 & 49.82 & 58.38 & -
\\ \midrule
\textsc{XAttn} & 72.49 & 73.84 & {57.77} & 47.52 & 57.59 & Yes
\\
\textsc{Sparge} & 65.08 & 66.86 & 47.09 & 40.43 & 49.37 & Yes
\\
\textsc{SlowFast} & 72.49 & 68.60 & 53.16 & 49.82 & 56.47 & Yes
\\ 
\textsc{StarAttn} & 74.07 & 72.67 & 57.28 & 50.53 & 58.26 & Yes
\\
\textsc{APB} & 75.13 & 73.84 & \textbf{58.01} & 50.18 & 59.16 & \textbf{No}
\\
\rowcolor{pink!20}
\textbf{\name} & \textbf{77.78} & \textbf{74.42} & {57.77} & \textbf{50.71} & \textbf{59.76} & \textbf{No}
\\
\bottomrule
\end{tabular}}

\caption{Accuracies (\%) of LongVideoBench. 
% Higher scores indicate better performance, 
``Overall'' stands for complete dataset's accuracy, and ``P. D.'' is the performance degredation compared with \textsc{FullAttn}. 
% The highest score in each column is marked in \textbf{bold}.
}
\label{tab:results-longvideobench}
\vspace{-10pt}
\end{table}

\subsection{Benchmarking Task Performance}

% \begin{figure*}[t]
% \begin{center}
% \raisebox{-\height}{\includegraphics[width=0.176\linewidth]{figures/speed-2b.pdf}}
% \raisebox{-\height}{\includegraphics[width=0.36\linewidth]{figures/speed-3b.pdf}}
% \raisebox{-\height}{\includegraphics[width=0.36\linewidth]{figures/speed-7b.pdf}}
% \end{center}
% \caption{The relative speedup of \name~and baselines compared to \textsc{FlashAttn}, evaluated across four input resolutions: 360p, 720p, 1080p, and 1440p. Note that \texttt{InterVL3-2B} resizes all inputs to 448$\times$448, so resolution has no impact on speedup.}
% \label{fig:speed}
% \end{figure*}

% \begin{figure*}[t]
% \begin{center}
% \includegraphics[width=0.196\linewidth]{figures/speed-2b.pdf}
% \includegraphics[width=0.397\linewidth]{figures/speed-3b.pdf}
% \includegraphics[width=0.397\linewidth]{figures/speed-7b.pdf}
% \end{center}
% \caption{The relative speedup of \name~and baselines compared to \textsc{FullAttn}, evaluated across four input resolutions: 360p, 720p, 1080p, and 1440p. Note that \texttt{InterVL3-2B} resizes all inputs to 448$\times$448, so resolution has no impact on speedup.
% }
% \label{fig:speed}
% \end{figure*}

In Table~\ref{tab:results-vnbench} and Table~\ref{tab:results-longvideobench}, to investigate (\textit{\textbf{Q1}}), we conduct evaluations of \name~and competitive baselines on both LongVideoBench and VNBench. 

Performance variations are more pronounced on synthetic videos compared with real-world inputs. As shown in Table~\ref{tab:results-vnbench}, all methods exhibit some accuracy degradation compared to \textsc{FullAttn}. Among them, \textsc{XAttn}, \textsc{APB}, and \name~exhibit better performance. 
\textsc{SlowFast} suffers significant accuracy drops, revealing the limitations of token pruning for complex and long retrieval tasks. \textsc{StarAttn} also underperforms as it cannot capture long-range dependencies. In contrast, \name~achieves the highest accuracy among all baselines. 

For real-world long videos (Table~\ref{tab:results-longvideobench}), \textsc{XAttn} and \textsc{SlowFast} show significant performance degradation across all models. While \textsc{StarAttn} outperforms sparse attention and token pruning methods, it remains less accurate than \textsc{FullAttn} on \texttt{InternVL3-2B} and \texttt{Qwen2.5VL-7B}. Both \textsc{APB} and \name~achieve near-lossless accuracy, with \name~surpassing \textsc{APB} on two models. 

Overall, \name~achieves better results among approximate attention methods for long-video understanding, surpassing all baselines.

\subsection{Benchmarking Inference Speed}

\begin{figure}[t]
\begin{center}
\includegraphics[width=1\linewidth]{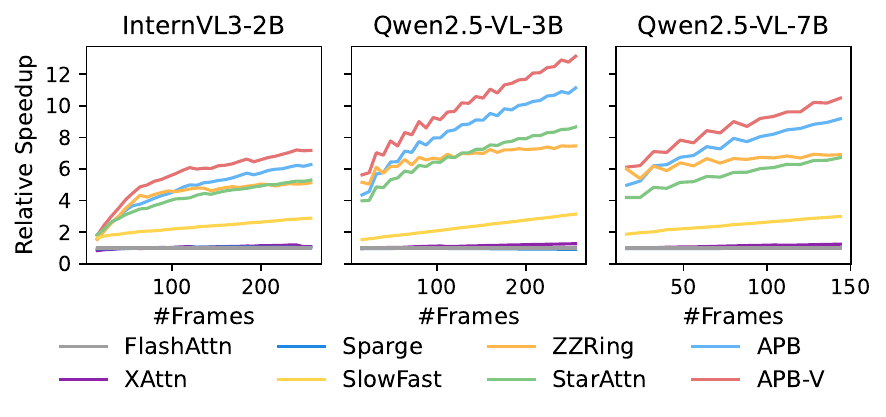}
\end{center}
% \vspace{-15pt}
\caption{The relative speedup of \name~and baselines compared to \textsc{FlashAttn} under various number of frames. We use 720p for \texttt{Qwen2.5-VL} models.}
\label{fig:speed-a}
% \vspace{-10pt}
\end{figure}

\begin{figure}[t]
\begin{center}
\includegraphics[width=1\linewidth]{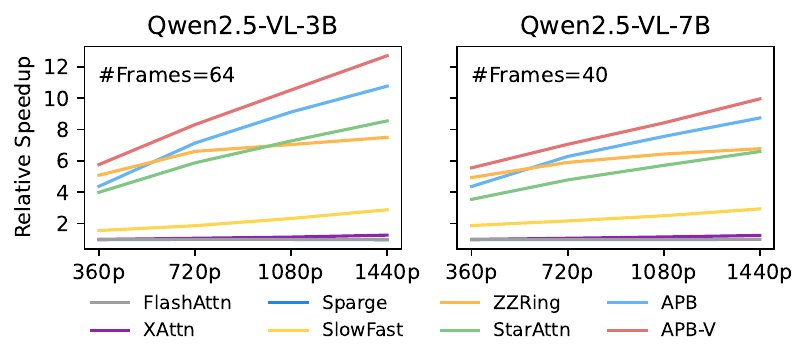}
\end{center}
% \vspace{-15pt}
\caption{The relative speedup of \name~and baselines compared to \textsc{FlashAttn} under various resolutions. }
\label{fig:speed-b}
% \vspace{-10pt}
\end{figure}

To address (\textit{\textbf{Q2}}), we evaluate the speedup of all methods relative to \textsc{FlashAttn} across various resolutions and video lengths, with results shown in Figure~\ref{fig:speed-a} and~\ref{fig:speed-b}. The speedup results show that: (1) sequence parallelism effectively enhances inference speed, where methods without sequence parallelism (\textsc{XAttn} and \textsc{SlowFast}) show significant limited speedup; (2) \name~outperforms all baselines across all scenarios regardless of base model, video resolution or video length; (3) \name~exhibits extraordinary speedup for high resolutions or long videos, achieving 12.72$\times$, 1.70$\times$, and 1.18$\times$ speedup compared with \textsc{FlashAttn}, \textsc{ZZRing}, and \textsc{APB}, respectively, when processing 64-frame 1440p videos on \texttt{Qwen2.5-VL-3B}.

\subsection{Ablation Studies}

To address (\textbf{\textit{Q3}}), we present an ablation study that highlights the contributions of each optimization, from both the algorithm and system perspectives.

\textbf{Block Composition.}
To analyze the impact of anchor block $\mathbf{A}$ and passing block $\mathbf{P}$, we conduct an ablation study on VNBench using the LMM \texttt{InternVL3-2B}. 
% The effect of each component is tested by removing them. 
Table~\ref{tab:abl-ap} demonstrates that: removing either component leads to significant performance degradation, with $\mathbf{P}$ particularly crucial for the overall accuracy. This validates the effectiveness of our design on applying an anchor block and a passing block for attention at each host.
\begin{table}[t]
% \vspace{-5pt}
\small
\centering
\setlength{\tabcolsep}{1.5mm}
\scalebox{0.9}{
\begin{tabular}{l|cccc}
\toprule
\multirow{2}{*}{Method} & \multicolumn{4}{c}{Counting} \\
& E-1 & E-2 & I & Avg. \\
\midrule
w/o $\mathbf{A}$ & 50.00 & 7.33 & 27.33 & 25.33 \\
w/o $\mathbf{P}$ & 21.33 & 5.33 & 25.33 & 17.33 \\
\rowcolor{pink!20}
\textbf{\name} & \textbf{52.00} & \textbf{9.33} & \textbf{28.67} & \textbf{30.00} \\
\bottomrule
\end{tabular}}
% \vspace{-5pt}
\caption{Ablation study on the anchor block $\mathbf{A}$ and passing block $\mathbf{P}$,
using \texttt{Qwen2.5VL-3B} as the tested LMM.}
% Experiments are conducted on the counting tasks of VNBench with the LMM \texttt{Qwen2.5VL-3B}.}
% \vspace{-10pt}
\label{tab:abl-ap}
\end{table}

\textbf{System Optimizations.}
We conduct an ablation study on \name's system optimizations to evaluate their effectiveness, focusing on four key components: (1) overlapping communication with computation in attention, (2) fused context-query forward pass, (3) ZigZag-style load balancing, and (4) visual encoding frame parallelism. We systematically disable each optimization and report the inference speed (req / s) in Table~\ref{tab:abl-sys}. We configure the anchor length as $l_a = {n\over64}$ and passing length as $l_p = {n\over128}$, using $H=8$ physical hosts. For ZigZag ablation, we revert to \textsc{APB}'s passing length ($l_p = n/64$). When the frame parallelism is removed, we encode all frames for each host. The ablation results demonstrate that all four system optimizations effectively accelerate the inference process. The fused context-query forward pass and frame parallelism yield the most significant speed improvements. When removing all optimizations, the system degrades to APB-like inference with all frames encoded on each host, remaining faster than \textsc{FlashAttn} but 4$\times$ slower than \name.
Overall, \name~exhibits the fastest inference speed.

\begin{table}[ht]
\small
\centering
\setlength{\tabcolsep}{1.5mm}
\scalebox{0.9}{
\begin{tabular}{l|cccccc}
\toprule
\multirow{2}{*}{Method} & \multicolumn{6}{c}{\#Frames} \\
& 16 & 24 & 32 & 40 & 48 & 56 \\
\midrule

\rowcolor{pink!20}
\textbf{\name} & \textbf{1.846} & \textbf{1.050} & \textbf{0.916} & \textbf{0.658} & \textbf{0.594} & \textbf{0.471} \\
\textsc{-O} & 1.827 & 1.049 & 0.911 & 0.656 & 0.593 & 0.470\\
\textsc{-O-F} & 1.646 & 0.980 & 0.854 & 0.624 & 0.565 & 0.450 \\
\textsc{-O-F-Z} & 1.618 & 0.947 & 0.813 & 0.590 & 0.523 & 0.415\\
\textsc{-O-F-Z-V} & 0.381 & 0.253 & 0.189 & 0.151 & 0.125 & 0.107\\
\textsc{FlashAttn} & 0.226 & 0.136 & 0.092 & 0.068 & 0.052 & 0.042 \\
\bottomrule
\end{tabular}}
% \vspace{-5pt}
\caption{Ablation on  system optimizations. Inference speed (req/s) of 1440p videos of various lengths are tested on \texttt{Qwen2.5VL-3B}. ``-'' indicates the baseline without certain optimization. ``O'', ``F'', ``Z'', and ``V'' represent overlapping communication-computation, fused context-query forward, ZigZag load-balancing strategy, and frame parallelism.}
\label{tab:abl-sys}
% \vspace{-10pt}
\end{table}

\textbf{Host Scalability.}
To evaluate the host scalability of \name~in terms of efficiency and performance, we conduct an ablation study comparing it with sequence-parallel baselines under different host counts $H \in \{2, 4, 6, 8\}$.  
Results are reported in Table~\ref{tab:abl-h-eff} and Table~\ref{tab:abl-h-per}.
Table~\ref{tab:abl-h-eff} shows that \name~achieves the fastest inference across all host counts for most frame numbers, whereas other baselines encounter severe efficiency bottlenecks as frame numbers grow. In terms of performance, Table~\ref{tab:abl-h-per} shows that \name~maintains stable task accuracy across different host counts.

\begin{table}[t]
\small
\centering
\setlength{\tabcolsep}{1.5mm}
\scalebox{0.9}{
\begin{tabular}{l|cccccc}
\toprule
\multirow{2}{*}{Method} & \multicolumn{6}{c}{\#Frames} \\
& 16 & 24 & 32 & 40 & 48 & 56 \\
\midrule
\multicolumn{7}{c}{$H=2$}\\
\midrule
\textsc{ZigZagRing} & \textbf{2.090} & 1.335 & 0.994 & 0.779 & 0.626 & 0.525\\
\textsc{StarAttn} & 1.401 & 0.908 & 0.666 & 0.519 & 0.416 & 0.345\\
\textsc{APB} & 1.788 & 1.208 & 0.909 & 0.726 & 0.592 & 0.503 \\
\rowcolor{pink!20} 
\textbf{\name} & 2.049 & \textbf{1.375} & \textbf{1.043} & \textbf{0.824} & \textbf{0.678} & \textbf{0.572} \\
\midrule
\multicolumn{7}{c}{$H=4$}\\
\midrule
\textsc{ZigZagRing} & \textbf{3.824} & 2.418 & 1.853 & 1.446 & 1.155 & 0.988 \\
\textsc{StarAttn} & 2.633 & 1.792 & 1.390 & 1.113 & 0.903 & 0.769 \\
\textsc{APB} & 3.131 & 2.227 & 1.740 & 1.436 & 1.176 & 1.008 \\
\rowcolor{pink!20} 
\textbf{\name} & 3.748 & \textbf{2.608} & \textbf{1.981} & \textbf{1.628} & \textbf{1.327} & \textbf{1.125}\\
\midrule
\multicolumn{7}{c}{$H=6$}\\
\midrule
\textsc{ZigZagRing} & 4.143 & 3.410 & 2.436 & 1.858 & 1.671 & 1.347\\
\textsc{StarAttn} & 3.157 & 2.605 & 1.942 & 1.537 & 1.379 & 1.146\\
\textsc{APB} & 3.505 & 3.070 & 2.273 & 1.867 & 1.702 & 1.437 \\
\rowcolor{pink!20} 
\textbf{\name} & \textbf{4.222} & \textbf{3.672} & \textbf{2.653} & \textbf{2.133} & \textbf{1.946} & \textbf{1.633} \\
\midrule
\multicolumn{7}{c}{$H=8$}\\
\midrule
\textsc{ZigZagRing} & 5.595 & 3.550 & 3.143 & 2.301 & 2.000 & 1.666\\
\textsc{StarAttn} & 4.485 & 2.888 & 2.571 & 1.975 & 1.762 & 1.493 \\
\textsc{APB} & 4.891 & 3.290 & 2.985 & 2.348 & 2.118 & 1.766\\
\rowcolor{pink!20} 
\textbf{\name} & \textbf{6.171} & \textbf{4.060} & \textbf{3.612} & \textbf{2.756} & \textbf{2.521} & \textbf{2.013}\\

\bottomrule
\end{tabular}}
% \vspace{-5pt}
\caption{Ablation across various host number ``H''. Inference speed (req/s) of 720p videos of various lengths are tested on \texttt{Qwen2.5VL-3B}.}
\label{tab:abl-h-eff}
% \vspace{-10pt}
\end{table}

\begin{table}[ht]
\small
\centering
\scalebox{0.9}{
\begin{tabular}{l|cccc}
\toprule
\multirow{2}{*}{Video Length} & \multicolumn{4}{c}{$H$} \\
& 2 & 4 & 6 & 8 \\
\midrule

8-15s & 68.78 & 69.84 & 70.37 & 68.25\\
15-60s & 69.77 & 70.93 & 71.51 & 70.93\\
180-600s & 51.94 & 52.91 & 53.64 & 53.88\\
90-3600s & 43.79 & 43.44 & 44.33 & 44.86\\
\midrule
Overall & 53.18 & 53.63 & 54.38 & 54.30\\
\bottomrule
\end{tabular}}
% \vspace{-5pt}
\caption{Performance under various host number $H$, where we select $H\in \{2, 4, 6, 8\}$. We evaluate LongVideoBench's performance using \texttt{Qwen2.5VL-3B}.}
\vspace{-5pt}
\label{tab:abl-h-per}
\end{table}

\begin{table}[ht]
\small
\centering
\scalebox{0.9}{
\begin{tabular}{l|cc}
\toprule
\multirow{3}{*}{Method} & \multicolumn{2}{c}{Host Setting}  \\
& (8) & \multicolumn{1}{c}{(4, 4)} \\
\midrule
\textsc{ZigZagRing} & 1.221 & 1.179 (-3.44\%)  \\
\textsc{StarAttn} & 0.959 & 0.939 (-2.09\%) \\
\textsc{APB} & 1.231 & 1.214 (-1.38\%)\\
\rowcolor{pink!20}
\textbf{\name} & \textbf{1.461} & \textbf{1.450} (\textbf{-0.75\%}) \\
\bottomrule
\end{tabular}}
% \vspace{-5pt}
\caption{Inference speed (req/s) across different communication media. “(8)” denotes 8 GPUs interconnected via NVLINK, while “(4, 4)” indicates two groups of 4 GPUs each, with intra-group NVLINK connections and inter-group communication over InfiniBand. We use \texttt{Qwen2.5-VL-7B} as the tested LMM.}
% \vspace{-5pt}
\label{tab:abl-nvlink}
\end{table}

Since \name~introduces communication, we further study the impact of different communication media on efficiency.  
As mainstream GPU clusters are typically connected via NVLINK or InfiniBand (IB), we use $H=8$ GPUs with NVLINK as the baseline, and then split them into two groups: NVLINK for intra-group links and IB for inter-group links, to measure the drop in inference speed. Thus, the IB link emerges as the bottleneck for the balanced \texttt{AllGather} operation.
As shown in Table~\ref{tab:abl-nvlink}, introducing the IB bottleneck degrades efficiency for all methods.  
\name~experiences the smallest drop and sustains stable inference speed, owing to its reduced communication design.

\begin{table}[h]
\small
\centering
\setlength{\tabcolsep}{1.5mm}
\scalebox{0.9}{
\begin{tabular}{l|ccccc}
\toprule
$l_p$ & 0 & 128 & 256 & 512 & No Comp \\
\midrule
Ordering-E (\%) & 12.67 & 34.00 & 37.33 & 40.00 & 40.67 \\
Throughput (req/s) & 5.67 & 5.63 & 4.78 & 4.39 & 4.26 \\
\bottomrule
\end{tabular}}
% \vspace{-5pt}
\caption{Hyperparameter analysis on $l_p$.}
\vspace{-10pt}
\label{tab:abl-lp}
\end{table}

\begin{table}[h]
\small
\centering
\setlength{\tabcolsep}{1.5mm}
\scalebox{0.9}{
\begin{tabular}{l|ccccc}
\toprule
$l_a$ & 0 & 128 & 256 & 512 & No Comp \\
\midrule
Ordering-E (\%) & 36.67 & 35.33 & 38.00 & 37.33 & 39.33 \\
Throughput (req/s) & 4.73 & 4.60 & 4.77 & 4.78 & 4.36 \\
\bottomrule
\end{tabular}}
% \vspace{-5pt}
\caption{Hyperparameter analysis on $l_a$.}
\vspace{-10pt}
\label{tab:abl-la}
\end{table}

\textbf{Hyperparameter Analysis.} we provide a detailed hyperparameter analysis on the anchor length $l_a$ and passing length $l_p$. The block length is determined by the input length and the number of hosts $H$.
We test \name~with different hyperparameters on the Ordering-E subset of VNBench, using InternVL3-2B. In our original setup, $l_p$ was set to $n/128$, which equals 256 in the VNBench experiment; $l_a=n/64$, which equals 512. This setting is consistent across all samples, since InternVL3-2B resizes input frames to 448×448 and we use 64 frames in the experiment.

For passing length $l_p$, we conduct experiments with $l_p \in {0, 128, 256, 512}$, as well as a ``no-compression'' setting. Note that due to implementation constraints in our modified \textsc{FlashAttn} kernel, $l_p$ must be a multiple of 128. Therefore, in the ``no-compression'' setting, we set $l_p$ to the largest possible value. The number of missing KVs is fewer than 256 per physical host (fewer than 128 on each virtual host). The results in Table~\ref{tab:abl-lp} show a 1.12$\times$ speedup when compressing $\mathbf{B}$ to $\mathbf{B}_c$ with $l_p = n/128$. When the compression rate is higher (i.e., $l_p = 0$ or $128$), we observe a notable degradation in model performance. Therefore, we choose $l_p = n/128$ as the default setting to strike a balance between efficiency and performance.

For anchor length $l_a$, we conduct experiments with $l_a\in \{0, 128, 256, 512, 1024\}$. The results in Table~\ref{tab:abl-la} show that the sensitivity of $l_a$ is lower than that of $l_p$. However, when $l_a$ is small (0 or 128), some performance degradation occurs. When $l_a$ is large (1024), the throughput decreases significantly. Therefore, we select $l_p = 512$.

\subsection{Case Study}
\label{sec:case}

\begin{figure*}[t]
\begin{center}
\includegraphics[width=\linewidth]{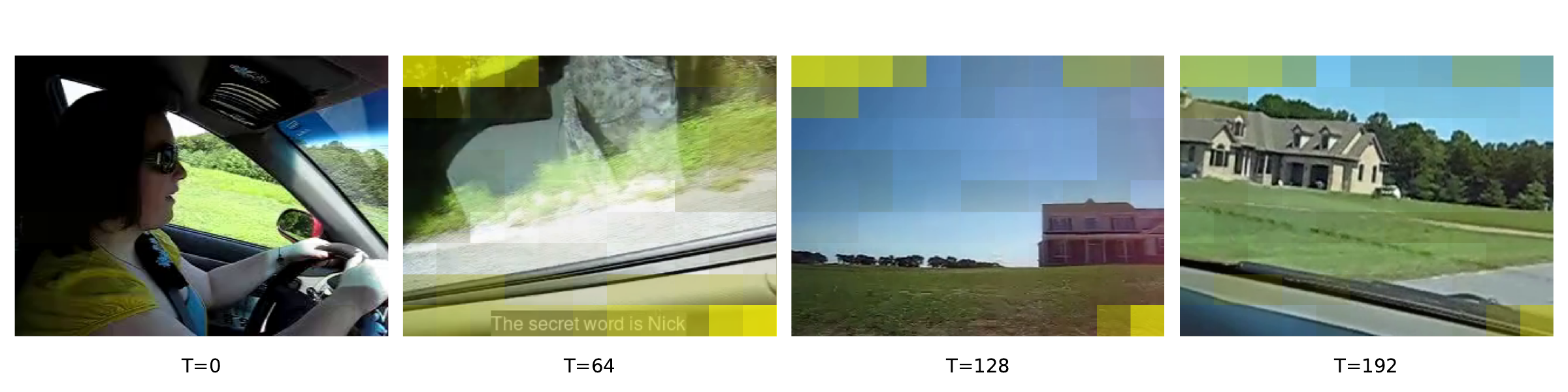}
\end{center}
\vspace{-10pt}
\caption{A case study from VNBench. Yellow intensity indicates how frequently a spatial position is selected into the passing block. Given the text query ``What is the secret word in this video?'', the correct answer is ``Nick''. The region corresponding to ``Nick'' exhibits noticeably higher yellow intensity than other regions.
}
\label{fig:case}
\vspace{-10pt}
\end{figure*}

To fully demonstrate how the approximate attention works in our proposed algorithm, we conduct a case study. We select one video from the Retrieval-E subset of VNBench. This video has 256 frames and is a first-person recording taken inside a car, capturing a group of friends on a road trip along a rural country road as shown in Figure~\ref{fig:case}. A subtitle reading “The secret word is Nick” is inserted from frames 63 to 67. The query for this video is ``What is the secret word in this video?''. We count how many times each video block is selected into the passing blocks (i.e., the essential KVs) and visualize this frequency as yellow intensity. As shown in Figure~\ref{fig:case}, the region containing the answer exhibits noticeably higher yellow intensity than other regions, indicating that it is selected into the passing blocks more frequently. Consequently, the relevant evidence of the given query remains consistently available to each host, enabling the model to produce the correct answer.

\section{Related Works}

Existing methods on optimizing the efficiency of long-video inference can be categorized into two major types: \textit{intrinsic attention optimization} and \textit{explicit input reduction}. 

\textbf{Intrinsic attention optimization.} 
% Increasing input sequence lengths intensifies the computational demands of Transformer-based models. 
% To mitigate this challenge, various optimizations target the attention mechanisms of these models. These optimizations can be achieved through KV cache-centric strategies, memory offloading, and approximate attention designs. 
To mitigate the challenge of intense computational demands caused by increasing input sequence lengths, various optimizations target the attention mechanisms of these models. These optimizations can be achieved through KV cache-centric strategies, memory offloading, and approximate attention designs. 
\textit{KV cache–centric strategies}~\citep{zhang2023h2o, li2024snapkv, xiao2024duoattention, liu2024kivi, huang2024locret, kim2024infinipot} focus on reducing the KV cache size to lower memory usage and boost inference throughput. \textit{Memory offloading techniques}~\citep{lee2024infinigen, xiao2024infllm, sun2024shadowkv} reduce both memory and computation by selectively loading only the most relevant cache blocks onto the GPU for attention computation. \textit{Approximate attention methods}~\citep{jiang2024minference, li2025mminference, xu2025xattention, zhang2025spargeattn, zhang2025sageattention, zhang2024sageattention2} compute only a subset of attention scores to lower computational cost while preserving model performance. Approximate attention methods can also be integrated with sequence parallelism~\citep{acharya2024starattention, huang2025apb} to further accelerate inference using multi-GPU systems. Recently, trainable approximate attention methods~\citep{yuan2025native, lu2025moba, gao2024seerattention, team2025minicpm4} have also shown potential in achieving both high performance and efficiency.

\textbf{Explicit input reduction.} Accelerating long-video inference can also be accomplished by reducing the number of input tokens before these tokens are processed by the visual encoder or the LLM backbone. \textit{LLM token reduction} methods reduce the number of input tokens before the LLM backbone to alleviate the computational burden. These techniques~\citep{xu2024slowfast, lee2024video, shi2025static, luo2025quota, wang2024dynamic} typically select the most important or merge video embeddings while minimizing performance loss. Other methods~\citep{xu2025slowfast, yao2025timechat,yu2025minicpm} perform token reduction-aware multimodal training. Although these methods effectively compress video tokens for the LLM, they fail to address the cost of visual encoding, which remains a non-negligible bottleneck in the long-video inference process. \textit{Video token reduction}~\citep{choudhury2024don, zhou2024video, zhou2023propainter, hao2024principles, jang2025efficient, ataiefard2024skipvit} aims to reduce the computational cost of visual encoding by exploiting temporal redundancy and spatial sparsity immediately after video patchification. While effective in accelerating both the visual encoder and the LLM backbone, such methods may suffer from notable performance degradation on complex videos, as they often overlook fine-grained details.

\section{Conclusion}
We propose \name, a sequence parallelism framework featuring an approximate attention mechanism for LMM long-video inference. 
% \name~applies frame parallelism for visual encoding and sequence parallelism-aware approximate attention for the LLM backbone. 
Through local KV cache compression and passing blocks, \name~simultaneously reduces communication and computation costs while maintaining long-range dependencies. 
Carefully crafted system optimizations ensure \name's efficiency.
Extensive evaluations demonstrate \name's superiority, achieving 12.72$\times$, 1.70$\times$, and 1.18$\times$ speedup over \textsc{FlashAttn}, \textsc{ZigZagRing}, and \textsc{APB}, respectively, without notable performance loss.

\section*{Limitations}

Since \name~primarily optimizes attention mechanisms, our benchmarks focus on decoder-only transformer-based LMMs.  
Other architectures, such as convolutional networks, are incompatible with the proposed methods.  
We also restrict \name~to multi-GPU inference, as on a single GPU it degenerates to \textsc{FlashAttn}.
\name~mainly focuses on accelerating long-video inference with a scalable number of GPUs under scenarios constrained by end-to-end time (i.e., time-to-first-token, TTFT). A wide range of real-world applications for long-video understanding align with these requirements, e.g., surveillance camera video processing, autonomous driving, etc. Running long-video inference on a single GPU is heavily limited by the total compute power of that GPU. For example, if a video is encoded into 512K tokens for an 8B LMM, the theoretical wall-time for prefilling such a sequence on a single NVIDIA A100 would exceed 5 minutes, which is unacceptable in many scenarios. Therefore, accelerating long-video inference with more compute power, i.e., more GPUs, is the only solution. Therefore, we restrict our discussion to scenarios where multiple GPUs are available.

\section*{Acknowledgments}

This work is supported by the  National Key Research and Development Program of China (2024YFB4505603) and National Natural Science Foundation of China (No. 62576186) and a grant from the Guoqiang Institute, Tsinghua University.

\bibliography{custom}
\newpage
\appendix

\section{Pseudocode of \name's Core Attention}
\label{sec:pseudocode}

\begin{algorithm}[h]
    \small
    \DontPrintSemicolon
    \caption{\name's Core Attention}
    \label{alg:infer}
    \KwIn{Host index $h$, Host number $H$;\\ Ancore Block $\{\mathbf{Q}_a, \mathbf{K}_a, \mathbf{V}_a\}$;\\ 
    Context Block 1 $\{\mathbf{Q}^{(h)}, \mathbf{K}^{(h)}, \mathbf{V}^{(h)}\}$;\\ 
    Context Block 2 $\{\mathbf{Q}^{(2H-1)}, \mathbf{K}^{(2H-1)}, \mathbf{V}^{(2H-1)}\}$;\\ 
    Query Block $\{\mathbf{Q}_{qr}, \mathbf{K}_{qr}, \mathbf{V}_{qr}\}$.}
    \KwOut{Attention score of anchor block $\mathbf{A}_a$,\\ context block $\mathbf{A}^{(h)}$, $\mathbf{A}^{(2H-1)}$, and query block $\mathbf{A}_{qr}$.}

    \color{blue}
    \tcp{{Identify Essential KVs}}
    \color{black}

    $\left[\mathbf{S}_{h},\mathbf{S}_{2H-1}\right] \leftarrow \mathbf{Q}_{qr}\left[\mathbf{K}^{(h)},\mathbf{K}^{(2H-1)}\right]^\top$\;

    \color{blue}
    \tcp{{Send Block 1's Essential KVs}}
    \color{black}

    $\mathbf{K}_c^{(h)}, \mathbf{V}_c^{(h)} \leftarrow (\mathbf{K}^{(h)}, \mathbf{V}^{(h)})\texttt{.gather}(\texttt{ArgTop}_{l_p}(\mathbf{S}_h))$\;

    \texttt{handle}$_1 \leftarrow $ \texttt{AsyncAllGather}($\mathbf{K}_c^{(h)}, \mathbf{V}_c^{(h)}$)\; 

    \color{blue}
    \tcp{{Send Block 2's Essential KVs}}
    \color{black}

    $\mathbf{K}_c^{(2H-1)}, \mathbf{V}_c^{(2H-1)} \leftarrow (\mathbf{K}^{(2H-1)}, \mathbf{V}^{(2H-1)})\texttt{.gather}(\texttt{ArgTop}_{l_p}(\mathbf{S}_{2H-1}))$\;

    \texttt{handle}$_2 \leftarrow $ \texttt{AsyncAllGather}($\mathbf{K}_c^{(2H-1)}, \mathbf{V}_c^{(2H-1)}$)\;
    
    \color{blue}
    \tcp{{Query Attention}}
    \color{black}

    $\mathbf{A}_{qr}^{(h)}, \texttt{lse}^{(h)} \leftarrow \texttt{QueryAttention}(
        \mathbf{Q}_{qr},$
        $[\mathbf{K}_a, \mathbf{K}^{(h, 2H-1)}, \mathbf{K}_{qr}],
        [\mathbf{V}_a, \mathbf{V}^{(h, 2H-1)}, \mathbf{V}_{qr}]
    )$\;
    \texttt{handle}$_3 \leftarrow $ \texttt{AsyncAllGather}($\mathbf{A}_{qr}^{(h)}, \texttt{lse}^{(h)}$)\;

    \color{blue}
    \tcp{{Anchor and Block 1's Attention}}
    \color{black}

    $\mathbf{K}_p^{(h)}, \mathbf{V}_p^{(h)} \leftarrow \texttt{MakePassingBlock}(\texttt{handle}_1)$

    $\mathbf{A}_a, \mathbf{A}^{(h)} \leftarrow \texttt{Attn}(
        [\mathbf{Q}_a, \mathbf{Q}^{(h)}],$
        $[\mathbf{K}_a, \mathbf{K}_p^{(h)}, \mathbf{K}^{(h)}],
        [\mathbf{V}_a, \mathbf{V}_{p}^{(h)}, \mathbf{V}^{(h)}]
    )$\;

    \color{blue}
    \tcp{{Block 2's Attention}}
    \color{black}

    $\mathbf{K}_p^{(2H-1)}, \mathbf{V}_p^{(2H-1)} \leftarrow \texttt{MakePassingBlock}(\texttt{handle}_2)$

    $\mathbf{A}^{(2H-1)} \leftarrow \texttt{Attn}(
        \mathbf{Q}^{(2H-1)},$
        $[\mathbf{K}_a, \mathbf{K}_p^{(2H-1)}, \mathbf{K}_p^{(2H-1)}],
        [\mathbf{V}_a, \mathbf{V}_p^{(2H-1)}, \mathbf{V}_p^{(2H-1)}]
    )$\;
    
    \color{blue}
    \tcp{{Merging Query Attention Result}}
    \color{black}

    $\mathbf{A}_{qr}\leftarrow\texttt{Merge}(\texttt{handle}_3)$\;
    
    \Return $\mathbf{A}_a, \mathbf{A}^{(h)}, \mathbf{A}^{(2H-1)}, \mathbf{A}_{qr}$\;
\end{algorithm}

\section{Details of System Optimizations}
\label{sec:sys_opt}

\subsection{Fused Context and Query Forward}

Previous methods, including \textsc{StarAttn}~\citep{acharya2024starattention} and \textsc{APB}~\citep{huang2025apb}, perform the prefill stage using two separate forward passes: one for the document context and another for the query. Since queries are typically short, a single forward pass dedicated to the query can become memory-bound and inefficient.
To address this issue, we fuse the prefill processes of both the context (video) and the query into a single forward pass, enabling faster prefill execution by reducing memory I/O of model's parameters. Apart from reading the model's parameters twice in existing methods, we concatenate the query block after each host's context block and conduct all linear projections together for both blocks to reduce redundant memory I/O. As described in the inference process section, we conduct an online \texttt{softmax} on query attention's partial result $\textbf{A}_{qr}^{(h)}$ and $\texttt{lse}^{(h)}$ at the end of each attention module to enable the query attention to be fully completed within the same forward pass as the video context.

\subsection{Approximate Attention Load Balancing}

\textsc{APB}~\citep{huang2025apb} introduces imbalanced computation across hosts, as the number of context blocks involved in the attention computation varies from host to host. This imbalance limits overall efficiency, since the attention computation time is ultimately bounded by the last host.
To address this issue, we develop a ZigZag-style load balancing strategy inspired by \textsc{RingFlashAttn}~\citep{zhu2024ringflash}, enabling more balanced \textsc{APB} computation across hosts.

Given $H$ physical hosts, we instantiate $2H$ virtual hosts and assign virtual hosts $h$ and $2H-1-h$ to physical host $h$. 
Since anchor blocks are identical, each host only holds one replica of the anchor block, and so is the query block.
In this setup, each host processes one anchor block of length $l_a$, two context blocks of length $l_b$, and a total of $2H - 1$ passing blocks of length $l_p$, ensuring an $h$-independent and equal amount of compute for all physical hosts. 
The amout of FLOPs of one attention computation on physical host $h$ is
\begin{align}
    2l_a^2d + 4l_b^2d + 4(2H-1)l_p l_b d,
\end{align}
which is independent of $h$ and identical across all physical hosts.

Since the anchor block introduces a significant amount of compute in the query attention, we also design a load balancing strategy specifically for this stage. As each physical host holds a full copy of the anchor block, we evenly divide the anchor block into $H$ slices. The query on physical host $h$ attends only to the $h$-th slice of the anchor block. 
The numerical accuracy is ensured by the online \texttt{softmax} technique.

\subsection{Overlapping Communication with Computation}

Existing methods such as \textsc{StarAttn}~\cite{acharya2024starattention} avoid communication across hosts and therefore cannot effectively capture long-term dependencies. In contrast, \textsc{APB}~\citep{huang2025apb} addresses long-term dependencies by introducing inter-host communication; however, this comes with additional overhead, as attention computation becomes tightly coupled with communication results.
To mitigate this issue, we design an overlapping strategy to eliminate GPU compute bubbles during communication by structuring the attention process in a two-stage manner. Specifically, for the $h$-th host, we first perform attention over the anchor block $\mathbf{B}_a$ and the local context block $\mathbf{B}^{(h)}$, followed by a second-stage attention over $\mathbf{B}^{(2H-1-h)}$ separately.
Since attention computation is highly compute-bound, dividing it into two stages introduces minimal overhead while allowing communication to be overlapped with useful computation.

When the attention mechanism begins, we first compute the multiplication $\mathbf{Q}_{qr}\mathbf{K}^{(h, 2H-1-h)\top}$ to obtain the importance scores, followed by gathering the compressed context blocks for virtual hosts $h$ and $2H-1-h$. Once the $\mathbf{B}_c^{(h)}$ and $\mathbf{B}_c^{(2H-1-h)}$ are generated, we perform \texttt{AllGather} communication for them while simultaneously computing the query attention.
Then, we perform the first-stage attention over the anchor block $\mathbf{B}_a$ and context block $\mathbf{B}^{(h)}$, leveraging the fact that the passing block for virtual host $h$ is received during query attention. Upon receiving the passing block from virtual host $2H-1-h$, we proceed with the second stage of attention for $\mathbf{B}_c^{(2H-1-h)}$.
The communication of the partial query attention results and \texttt{lse} takes place immediately after the communication of the two passing blocks, ensuring that the merging of the query attention can proceed without delay following the second stage.
As a result, all communication steps can be effectively overlapped with computation, without any compute waiting bubbles. The detailed overlapping routine is illustrated in Figure~\ref{fig:overlap}.

\section{Baselines and More Related Works}

We first introduce our baselines, followed by a brief discussion on the potentially related methods.

\begin{table*}[t]
\small
\centering
\setlength{\tabcolsep}{1.5mm}
\scalebox{0.9}{
\begin{tabular}{l|cccc|cccc|cccc|cc}
\toprule
\multirow{2}{*}{Method} & \multicolumn{4}{c|}{Retrieval} & \multicolumn{4}{c|}{Ordering} & \multicolumn{4}{c|}{Counting} & \multirow{2}{*}{\makecell{Overall\\(\%) }} & \multirow{2}{*}{\makecell{Thru.\\ (req/s)}}\\

 & E & I-1 & I-2 & Avg. & E & I-1 & I-2 & Avg. & E-1 & E-2 & I & Avg. &\\
\midrule
\multicolumn{14}{c}{\texttt{InternVL3-2B}} \\ \midrule
\textsc{FullAttn} & 90.00 & 90.67 & 36.00 & 72.22 & 64.67 & 24.00 & 24.67 & 37.78 & 40.67 & 4.67 & 28.67 & 24.67 & 44.89 & 1.40
\\ \midrule
\textsc{FastV} & 84.00 & 90.00 & \textbf{39.33} & \textbf{71.11} & 27.33 & 8.67 & 16.00 & 17.33 & 31.33 & \textbf{5.33} & 24.67 & 20.44 & 36.30 & 1.89
\\
\rowcolor{pink!20}
\textbf{\name} & \textbf{90.67} & \textbf{89.33} & 32.67 & 70.89 & \textbf{64.00} & \textbf{22.00} & \textbf{21.33} & \textbf{35.78} & \textbf{37.33} & \textbf{5.33} & \textbf{26.67} & \textbf{23.11} & \textbf{43.26} & \textbf{4.78}
\\
\bottomrule
\end{tabular}}

\caption{Comparing with \textsc{FastV} on VNBench. 
% Higher scores indicate better performance, and ``Overall'' stands for the overall accuracy on the complete dataset. 
``E'' and ``I'' represent the edited and inserted data subset.
% The highest score in each column is marked in \textbf{bold}.
}
\label{tab:results-vnbench-fastv}
\end{table*}

\begin{table*}[t]
\small
\centering
\setlength{\tabcolsep}{1.5mm}
\scalebox{0.9}{
\begin{tabular}{l|ccccccccccc|cc}
\toprule
Method & SG1 & SG2 & MK1 & MK2 & MV & MQ & VT & CWE & FWE & QA1 & QA2 & Avg. & Thru. (tok/s)\\
\midrule
\textsc{FullAttn} & 100.00 & 100.00 & 90.00 & 90.00 & 90.00 & 100.00 & 64.00 & 66.00 & 65.00 & 75.00 & 40.00 & 80.00 & 4314.89
\\
\textsc{APB} & \textbf{100.00} & \textbf{100.00} & 80.00 & 85.00 & 93.75 & 98.75 & 64.00 & 69.00 & 71.67 & \textbf{70.00} & \textbf{30.00} & 78.29 & 41704.68
\\
\rowcolor{pink!20}
\textbf{\name} & \textbf{100.00} & \textbf{100.00} & \textbf{90.00} & \textbf{90.00} & 88.75 & \textbf{100.00} & \textbf{73.00} & \textbf{71.00} & \textbf{75.00} & \textbf{70.00} & \textbf{30.00} & \textbf{80.70} & \textbf{60376.24}
\\
\bottomrule
\end{tabular}}

\caption{Comparing \name~with \textsc{APB} on RULER. 
% Higher scores indicate better performance, and ``Overall'' stands for the overall accuracy on the complete dataset. 
% ``E'' and ``I'' represent the edited and inserted data subset.
% The highest score in each column is marked in \textbf{bold}.
}
\label{tab:results-ruler}
\end{table*}

\subsection{Baselines}

\textbf{\textsc{FlashAttn}.} This is an accurate attention implementation without sequence parallelism. Since the attention computation must be performed on a single GPU, only data parallelism or pipeline parallelism can be applied during deployment. However, these forms of parallelism do not reduce the TTFT (time-to-first-token) for long-video inference requests.

\textbf{\textsc{SpargeAttn} and \textsc{XAttn}.} These two methods introduce sparse attention with different sparsity patterns but do not incorporate sequence parallelism. They reduce computation by eliminating redundant attention operations that naturally arise from sparse attention scores, and thus can speed up long-video inference to some extent. However, without sequence parallelism, the computational capability of a single GPU remains insufficient to achieve acceptable TTFT.

\textbf{\textsc{SlowFast-LLaVA}.} This method introduces sparsity by pruning video embeddings. The attention module in the LLM backbone remains unchanged, and the speedup is mainly attributed to the reduced number of input video embeddings. However, removing embeddings introduces significant information loss, as the pruned embeddings cannot be recovered, resulting in notable performance degradation.

\textbf{\textsc{ZigZagRing}.} \textsc{ZigZagRing} serves as a vanilla baseline of sequence parallelism based on ring-style communication. The input context is evenly partitioned across hosts, and a zigzag load-balancing strategy is adopted to balance computation and communication during attention evaluation. For $H$ hosts, $H-1$ communication rounds are required, where each virtual host $h$ receives partial attention results from $h-1$ and sends to $h+1$. The outputs are aggregated using an online softmax. Because \textsc{ZigZagRing} does not introduce attention sparsity, it still suffers from dense computation when processing long-video inputs.

\textbf{\textsc{StarAttn}.} \textsc{StarAttn} is an early approach that integrates sparsity into sequence parallelism. Each host attends only to the anchor block (the beginning of the sequence) and its local block, with no communication during document prefill. Query prefill uses online softmax to merge partial results across hosts. \textsc{StarAttn} is fast and scales well with the number of hosts, but it cannot capture long-range dependencies due to its strictly local sparse attention pattern. In contrast, APB-V maintains long-context dependencies through training-free passing block construction and further improves efficiency through system-level optimizations.

\textbf{\textsc{APB}.} \textsc{APB} is proposed to mitigate the performance degradation observed in StarAttn. It incorporates passing blocks into each attention computation and selects essential KVs using a trained retaining head. Unlike \textsc{APB}, APB-V is completely training-free. Moreover, APB still requires two separate forward passes for the document and the query, and its lack of load-balancing strategy limits efficiency. With carefully designed system optimizations, APB-V achieves significantly higher speed than \textsc{APB}.

\subsection{Potentially Related Methods}

\textbf{\textsc{FastV}.} \textsc{FastV}~\citep{chen2024image} is a token-pruning method that reduces the number of hidden states after a certain layer of the LLM backbone. Because some hidden states are discarded to improve efficiency, substantial information is lost and the removed content cannot be recovered. In addition, the original \textsc{FastV} uses an H2O-like~\citep{zhang2023h2o} pruning function that requires access to full attention scores. This design is incompatible with FlashAttn, forcing \textsc{FastV} to fall back to vanilla matrix multiplication and resulting in high GPU memory usage (a limitation shared with H2O). For this reason, we replace it with a SnapKV-like pruning function in the experiments above, otherwise there would be out-of-memory errors. 

\textbf{\textsc{LongVU}.} \textsc{LongVU}~\citep{shen2024longvu} introduces frame reduction, feature fusion, and cross-modal video token pruning techniques to reduce computational cost when processing long video inputs. However, it requires large-scale training and incorporates prior knowledge from \textsc{DINOv2}~\citep{oquab2023dinov2}, making it unsuitable as a plug-and-play method for arbitrary pretrained LMMs. Another potential weakness compared with APB-V is that \textsc{LongVU} removes video tokens during the prefill stage, which may harm multi-turn QA performance. Once irrelevant tokens are pruned in the first turn, they cannot be retrieved in subsequent dialogue.

\textbf{\textbf{LlamaVID}.} \textsc{LlamaVID}~\citep{li2024llamavid} is an LMM specifically designed for high compression ratios in long-video processing, encoding each frame into only two tokens for the LLM backbone. This design requires large-scale multimodal pretraining and a carefully crafted pipeline for long-video understanding. In current LMMs, the information flow is typically simple: a ViT encodes long videos into embeddings, a connector compresses them into a compact representation, and the LLM processes them. Because aggressive compression is not always used during pretraining, a plug-and-play method (such as APB-V) offers a more practical solution that works broadly with existing models.

\section{Supplementary Experiments}

\subsection{Comparing with \textsc{FastV}}

\textsc{FastV} is an efficient inference method designed for LMMs where a \textsc{SnapKV}-like~\citep{li2024snapkv} pruning function is used to reduce model's KV cache. We test \textsc{FastV} on VNBench using \texttt{InternVL3-2B}, with $K$ set to 2 and $R$ set to 0.5. For \name~we set the host number $H$ to 8. The results in Table~\ref{tab:results-vnbench-fastv} show that \textsc{FastV} can accelerate long-video understanding by 1.35$\times$, but at the cost of a substantial performance degredation. In contrast, \name~achieves both higher performance and greater efficiency at the same time.

\subsection{Extending to Long-Context NLP Tasks}

\name~can be applied to any Transformer-based decoder-only model. To demonstrate its transferability, we evaluate \name~on long-context NLP tasks. We compare \name, \textsc{APB}, and \textsc{FullAttn} on RULER~\citep{hsieh2024ruler} using \texttt{Llama-3.1-8B-Instruct}~\citep{grattafiori2024llama}, and the results are listed below. We run 20 entries on the following tasks: Single-NIAH-1/2, MultiKey-NIAH-1/2, MultiValue-NIAH, MultiQuery-NIAH, VT, CWE, FWE, and QA1/2, using 8 GPUs. The input length is set to 128K. We also report the prefill throughput for reference. \textsc{FullAttn} is implemented in \textsc{FlashAttn} and is executed on a single GPU.
The results in Table~\ref{tab:results-ruler} show that \name~is also able to handle long-context NLP tasks. \name~achieves 13.99$\times$ and 1.45$\times$ speedups compared with \textsc{FullAttn} and APB, respectively.

\end{document}